\documentclass[sigconf]{acmart}

\usepackage{subfigure}
\usepackage{multirow}
\usepackage{stfloats}
\usepackage{pifont}

\copyrightyear{2025}
\acmYear{2025}
\setcopyright{acmlicensed}
\acmConference[CIKM '25] {Proceedings of the 34th ACM International Conference on Information and Knowledge Management}{ November 10--14, 2025}{Seoul, Republic of Korea.}
\acmBooktitle{Proceedings of the 34th ACM International Conference on Information and Knowledge Management (CIKM '25), November 10--14, 2025, Seoul, Republic of Korea}
\acmISBN{979-8-4007-2040-6/2025/11}
\acmDOI{10.1145/3746252.3761090}

\begin{document}

\title{DiRW: Path-Aware Digraph Learning for Heterophily}

\author{Daohan Su}
\email{dhsu@bit.edu.cn}
\orcid{0009-0003-1627-9608}
\affiliation{
  \institution{Beijing Institute of Technology}
  \city{Beijing}
  \country{China}}

\author{Xunkai Li}
\email{cs.xunkai.li@gmail.com}
\orcid{0000-0002-1230-7603}
\affiliation{
  \institution{Beijing Institute of Technology}
  \city{Beijing}
  \country{China}}

\author{Zhenjun Li}
\orcid{0000-0003-4176-8879}
\email{lizhenjun@szsit.edu.cn}
\affiliation{
  \institution{Shenzhen Institute of Technology}
  \city{Shenzhen}
  \country{China}}

\author{Yinping Liao}
\orcid{0009-0000-9661-0759}
\email{liaoyinping@szcp.edu.cn}
\affiliation{
  \institution{Shenzhen City Polytechnic}
  \city{Shenzhen}
  \country{China}}

\author{Rong-Hua Li}
\email{lironghuabit@126.com}
\orcid{0000-0002-3105-5325}
\affiliation{
  \institution{Beijing Institute of Technology}
  \city{Beijing}
  \country{China}}

\author{Guoren Wang}
\email{wanggrbit@gmail.com}
\orcid{0000-0002-0181-8379}
\affiliation{
  \institution{Beijing Institute of Technology}
  \city{Beijing}
  \country{China}}

\renewcommand{\shortauthors}{Daohan Su et al.}

\begin{abstract}
    Recently, graph neural network (GNN) has emerged as a powerful representation learning tool for graph-structured data.
    However, most approaches are tailored for undirected graphs, neglecting the abundant information in the edges of directed graphs (digraphs).
    In fact, digraphs are widely applied in the real world and confirmed to address heterophily challenges.
    Despite recent advancements, existing spatial- and spectral-based DiGNNs have limitations due to their complex learning mechanisms and reliance on high-quality topology, resulting in low efficiency and unstable performance.
    To address these issues, we propose \underline{Di}rected \underline{R}andom \underline{W}alk (DiRW), a plug-and-play strategy for most spatial-based DiGNNs and also an innovative model which offers a new digraph learning paradigm.
    Specifically, it utilizes a direction-aware path sampler optimized from the perspectives of walk probability, length, and number in a weight-free manner by considering node profiles and topologies.
    Building upon this, DiRW incorporates a node-wise learnable path aggregator for generalized node representations. 
    Extensive experiments on 9 datasets demonstrate that DiRW:
    \ding{192} enhances most spatial-based methods as a plug-and-play strategy;
    \ding{193} achieves SOTA performance as a new digraph learning paradigm.
    The source code and data are available at \href{https://github.com/dhsiuu/DiRW}{https://github.com/dhsiuu/DiRW}.
\end{abstract}




\maketitle

\section{Introduction}
\label{sec: Introduction}
    Graph neural networks (GNNs) have been widely used across node-~\cite{wu2019sgc, wang2020gcnlpa, li2024_atp}, link-~\cite{Zhang18link_prediction1,tan2023link_prediction4,li2024lightdic}, and graph-level tasks~\cite{liang2023hetdag, luo2024dagformer, thost2021dagnn} and achieve satisfactory performance. 
    Therefore, this graph-based deep learning technique holds great potential for applications, such as recommendation~\cite{su2024dcl, yang2023app_gnn_rec2}, financial analysis~\cite{balmaseda2023app_gnn_fina1, hyun2023app_gnn_fina2, qiu2023app_gnn_fina3}, and healthcare~\cite{pfeifer2022fedapp_gnn_dis1, ahmed2022fedapp_gnn_dis2}. 
    Despite their effectiveness, existing methods often overlook edge direction in natural graphs, leading to inevitable information loss and limited performance upper bound.

    Compared to undirected representations, digraphs are crucial for modeling complex real-world topologies (e.g., web flow monitoring and bioinformatics), capturing more intricate node relationships.
    Additionally, the recently proposed A2DUG~\cite{maekawa2023a2dug}, Dir-GNN~\cite{dirgnn_rossi_2023} and ADPA~\cite{sun2024breaking} reveal a key insight: \textit{Edge direction offers a new perspective for addressing the topological heterophily challenges that plague graph learning}.
    Despite growing attention, DiGNNs are still in their infancy and face the following inherent limitations:
    \ding{192} Spatial-based methods often stack multiple convolution layers with separate learnable parameters for out-edges and in-edges, resulting in over-smoothing concerns~\cite{yan2021ggcn} and high computational costs~\cite{he2022dimpa,kollias2022nste,zhou2022dhypr}.
    \ding{193} Spectral-based methods heavily rely on high-quality directed topology~\cite{zhang2021magnet, tong2020digcn}, and without this, extreme eigenvalues inevitably lead to sub-optimal performance.
    Strict theoretical assumptions also limit practical deployment in complex scenarios.
    Therefore, developing a more efficient paradigm for digraph learning is urgent.

    To enhance the usability of DiGNNs, this paper focuses on spatial-based methods and proposes a novel (directed) path-based learning mechanism (Di)PathGNNs.
    As we all know, the entanglement of homophily and heterophily, where connected nodes exhibit intricate feature distributions and labels, has recently posed a significant challenge~\cite{ma2021hete_gnn_survey1, luan2022hete_gnn_survey2, zheng2022hete_gnn_survey3, platonov2023hete_gnn_survey4}. 
    Researchers strive to achieve robust learning within this complex topology.
    In this context, compared to the traditional message-passing (i.e., neighbor expansion with strict spatial symmetry that disregards edge direction), we highlight the following advantages of DiRW to further clarify the motivation of our study:
    \ding{182} \textbf{Edge Direction and Node Order.} 
    The core of DiPathGNNs lies in performing well-designed random walks for each node, treating the paths as node-wise sequences. 
    \textit{Advantage: DiRW fully considers edge direction and preserves the order of nodes within the walking paths, which is crucial for capturing structural insights.}
    \ding{183} \textbf{Adaptive Expansion of Node Receptive Fields.}  
    DiPathGNNs adaptively extend hop-based neighbors to path-based neighbors by considering the characteristics of random walk, incorporating more homophilous signals. 
    \textit{Advantage: DiRW ensures message aggregation among nodes with the same label, equivalent to data augmentation and highlighting label-specific positive signals for prediction.}
    \ding{184} \textbf{Path-based Message Aggregation.}
    Based on the above homophily-aware paths, DiPathGNNs aggregate these messages for predicting each node through a learnable mechanism.
    \textit{Advantage: DiRW facilitates the dense aggregation of direction-aware homophilous information, thus enhancing node prediction.}
    
\begin{figure}[t]
\centering
\includegraphics[width=\linewidth]{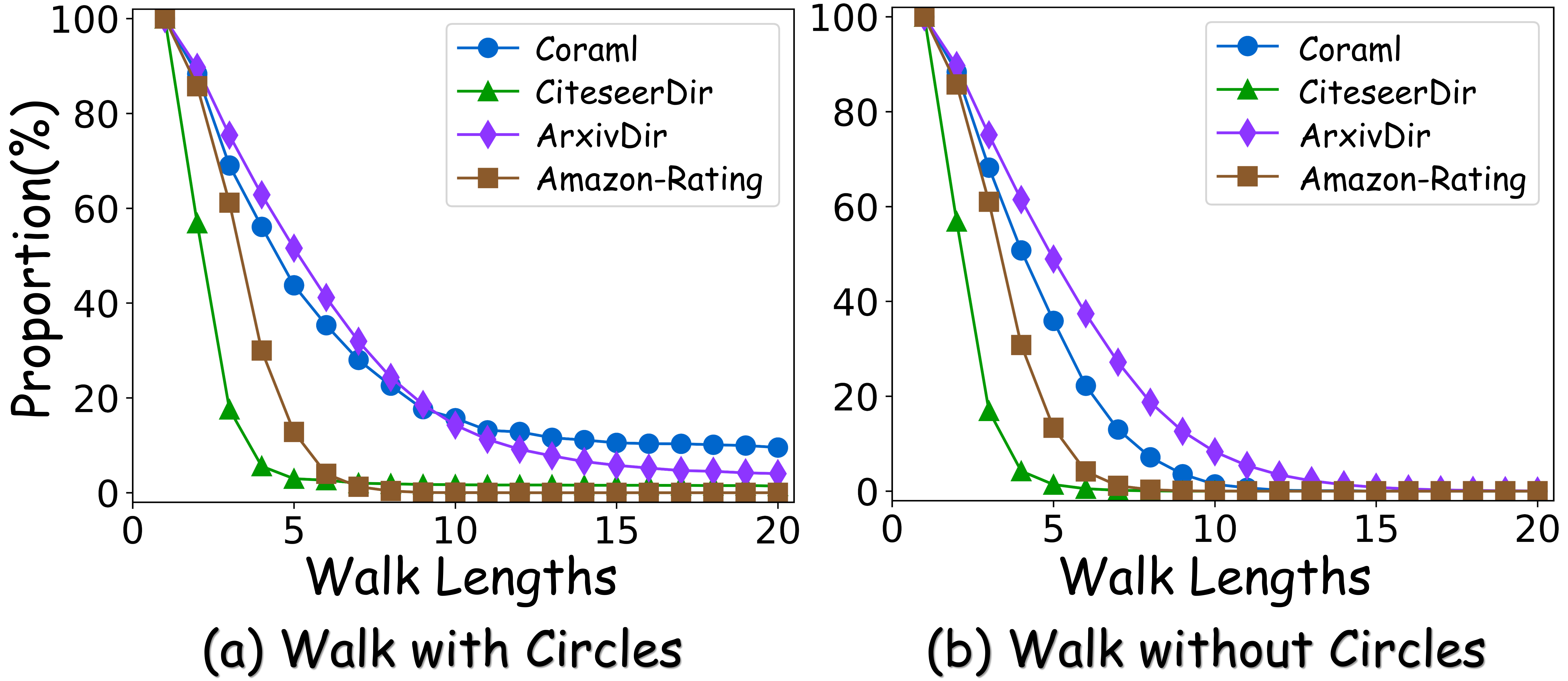}
\vspace{-6mm}
\caption{Interruption issue of DiSRW on digraphs.}
\label{fig: empirical_1}
\vspace{-4mm}
\end{figure}

\begin{figure}[t]
\centering
\includegraphics[width=\linewidth]{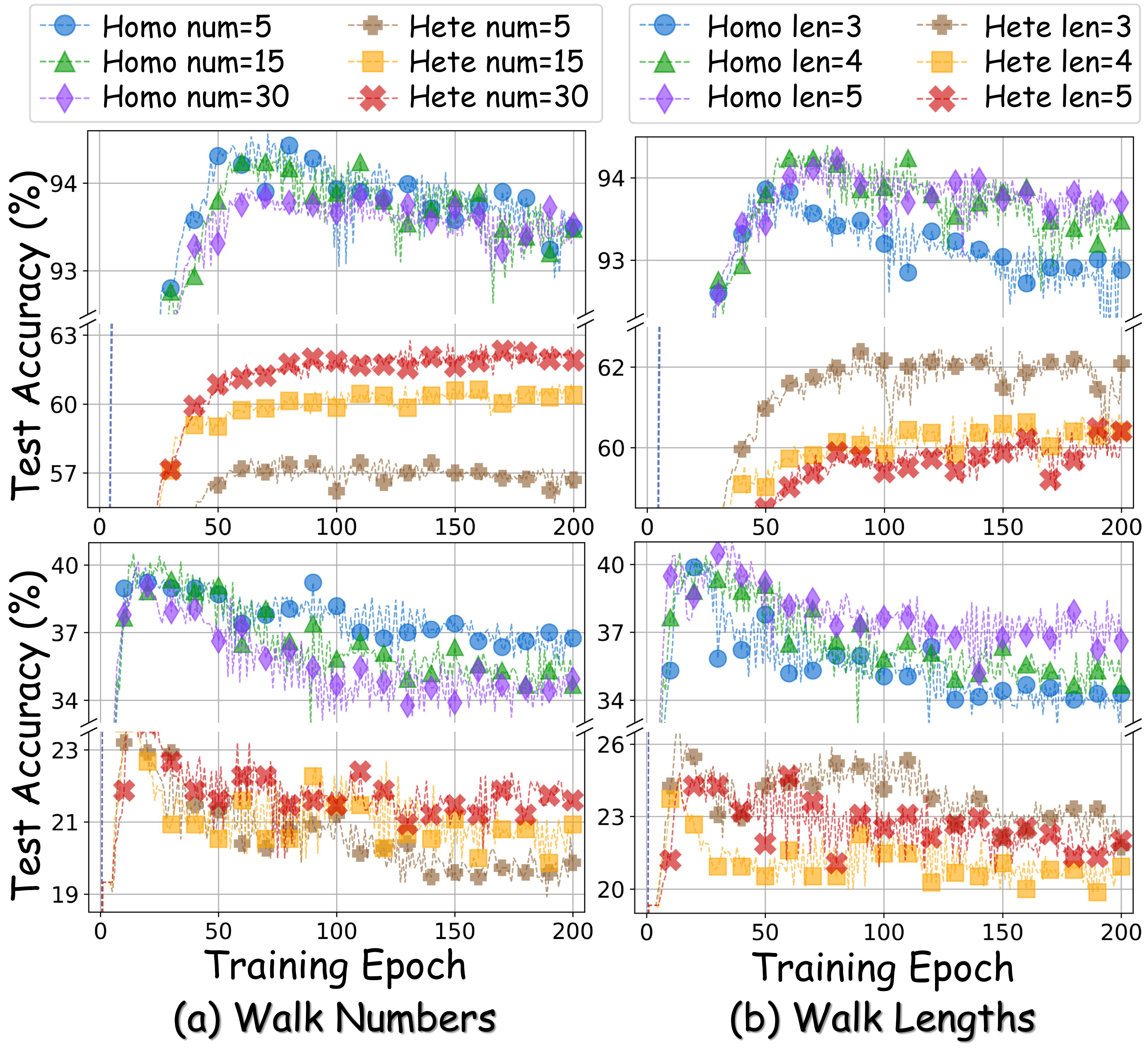}
\vspace{-6mm}
\caption{Performance of SRW on digraphs WikiCS (Homophilous, Upper) and Actor (Heterophilous, Lower).}
\label{fig: empirical_2}
\vspace{-4mm}
\end{figure}

    Despite recent advancements in PathGNNs~\cite{ijcai22_rawgnn, xie2023pathmlp, sun2022beyond}, their walk strategies are tailored for undirected graphs, lacking generalizability. 
    Furthermore, the complex relationships inherent in directed topology pose unique challenges to the naive random walks, necessitating further investigations to develop fine-grained walk rules.
    To further elucidate, we present empirical analyses in Fig.~\ref{fig: empirical_1}-\ref{fig: empirical_2} that demonstrate these limitations and offer key insights.

\ding{117} \textbf{Limitation 1: Neglect of edge directions.} 
    PathGNNs often fail to account for the edge direction. 
    Utilizing directed simple random walks (DiSRW) on digraphs inevitably leads to interruptions when encountering nodes lack outgoing edges.
    To investigate and visualize this limitation, we employ DiSRW and initiate a walk from each node on four digraphs.
    As the walk length increases, we record the proportion of complete paths out of the total sequences, as depicted in Fig.~\ref{fig: empirical_1}(a). 
    To mitigate the impact of cycles within the digraph, we also devise an DiSRW that excludes cycles and conduct the same experiment, as shown in Fig.~\ref{fig: empirical_1}(b).

\ding{80} \textbf{Key Insight 1:} 
    \textit{Walking strictly along the edge directions leads to severe walk interruption problems.} 
    The non-strongly connected nature of digraphs causes most walks to show a sharp decline in complete paths after just five steps, indicating that they fail to gather extensive information beyond the immediate neighborhood of the starting node.
    Additionally, removing the influence of cycles results in an even greater decline in uninterrupted sequences.

\ding{117} \textbf{Limitation 2: Coarse-grained walk strategies.}
    PathGNNs treat walk number and length as hyperparameters, uniformly applying them across all nodes. 
    This one-size-fits-all approach overlooks the distinct contexts of nodes within complex topology.
    To further investigate this, we conducted an in-depth analysis of the effects of varying walk numbers (Fig.~\ref{fig: empirical_2}(a)) and walk lengths (Fig.~\ref{fig: empirical_2}(b)) on node classification.
    We utilize undirected SRW as the sampling strategy, and utilize two MLPs to generate path embeddings from the nodes along the sampled paths and aggregate node embeddings from multiple paths associated with the same node.
    Node homophily is calculated as the ratio of the number of first-order neighboring nodes sharing the same label to the total number of first-order neighbors, with the top 50\% of nodes by homophily scores classified as homophilous nodes and the remains as heterophilous nodes.

\ding{80} \textbf{Key Insight 2:}
    \textit{Higher walk numbers facilitate heterophilous nodes, while longer walk lengths benefit homophilous nodes.} 
    Increasing walk numbers enhances predictions for heterophilous nodes by enabling a comprehensive exploration of their complex topologies.
    In contrast, homophilous nodes can be accurately represented with fewer walks due to their uniform neighborhood structure.
    Excessive walks in such case raise concerns regarding over-smoothing and time and space complexity.
    Regarding walk length, heterophilous nodes are often surrounded by noisy environments, where longer walks risk incorporating irrelevant information.
    Conversely, homophilous nodes benefit from longer walks as they gather a broader range of relevant information, offering a clearer surrounding views.
    
    Inspired by the above key insights, we propose \underline{Di}rected \underline{R}andom \underline{W}alk (DiRW), which can be viewed as a plug-and-play directed walk strategy for PathGNNs or a new digraph learning paradigm.
    Specifically, DiRW first designs a direction-aware walk strategy to identify potential neighbor relationships (motivated by \textbf{Key Insight 1}).
    It then fine-tunes walk numbers and lengths on a node-adaptive basis (motivated by \textbf{Key Insight 2}). 
    Finally, DiRW employs a node-wise learnable path aggregator to represent node embeddings, breaking the constraints of message-passing mechanisms.

\textbf{Our contributions.} 
    \ding{192} \textit{\underline{Novel Perspective}.} 
    This paper is the first to introduce DiPathGNN and highlight its advantages, emphasizing the need for fine-grained walking strategies in digraphs through valuable empirical studies.
    \ding{193} \textit{\underline{Plug-and-play Strategy}.} 
    We present DiRW, which customizes walk probabilities, lengths, and numbers for each node, seamlessly integrating with any DiPathGNN to enhance performance.
    \ding{194} \textit{\underline{New DiGNN}.} 
    DiRW also serves as a new learning architecture for digraphs, featuring a pre-processed optimized path sampler and a well-designed node-wise learnable path aggregator.
    \ding{195} \textit{\underline{SOTA Performance}.} 
    Experiments demonstrate that DiRW improves node classification by $2.81\%$ as a plug-and-play strategy and enhances link prediction by $0.82\%$ as a novel DiGNN.

\section{Preliminaries}
\label{sec: Preliminaries}

\subsection{Notation and Problem Formulation}
\label{sec: Notation}
    We consider a digraph $\mathcal{G}=(\mathcal{V},\mathcal{E})$ with $|\mathcal{V}|=n$ nodes and $|\mathcal{E}|=m$ edges, where each node has a feature of size $f$ and a one-hot label of size $c$.
    The feature and label matrices represented as $\mathbf{X} \in \mathbb{R}^{n \times f}$ and $\mathbf{Y} \in \mathbb{R}^{n \times c}$. The $\mathcal{G}$ is described by an asymmetrical adjacency matrix $\mathbf{A}$, where $\mathbf{A}(u, v) = 1$ if $(u, v) \in \mathcal{E}$, and $0$ otherwise.
    The goal of the semi-supervised node classification is to predict the labels for unlabeled nodes with the supervision of labeled nodes.

\subsection{Directed Graph Neural Networks}
\label{sec: Directed Graph Neural Networks}

\subsubsection{Spatial-based Methods}
\label{sec: Spatial-based Methods}
    To capture the asymmetric topology of digraphs, some spatial-based methods follow the strict symmetric message-passing paradigm in the undirected setting~\cite{huang2020cands, frasca2020sign, xu2018jknet}.
    However, it is crucial to account for the directionality of the edges when aggregating messages.
    Specifically, for the current node $u \in \mathcal{V}$, the model learns weights to combine the representations of its out-neighbors $(u\rightarrow v)$ and in-neighbors $(v\rightarrow u)$ independently:
    \begin{equation}
        \begin{aligned}
        \label{eq: Prevalent directed Message Passing}
        \mathbf{H}_{u,\rightarrow}^{(l)}=&\operatorname{Agg}\left(\mathbf{W}_{\rightarrow}^{(l)},\operatorname{Prop}\left(\mathbf{X}_u^{(l-1)},\left\{\mathbf{X}_v^{(l-1)},\forall (u,v)\in\mathcal{E}\right\}\right)\right),\\
        \mathbf{H}_{u,\leftarrow}^{(l)}=&\operatorname{Agg}\left(\mathbf{W}_{\leftarrow}^{(l)},\operatorname{Prop}\left(\mathbf{X}_u^{(l-1)},\left\{\mathbf{X}_v^{(l-1)},\forall (v,u)\in\mathcal{E}\right\}\right)\right),\\
        &\;\;\;\mathbf{X}_u^{(l)} = \operatorname{Com}\left(\mathbf{W}^{(l)}, \mathbf{X}_u^{(l-1)}, \mathbf{H}_{u,\rightarrow}^{(l)}, \mathbf{H}_{u,\leftarrow}^{(l)}\right),
        \end{aligned}
    \end{equation}
    where the Propagation function $\operatorname{Prop}\left(\cdot\right)$ gathers and distributes feature from neighbors, and the Aggregation function $\operatorname{Agg}\left(\cdot\right)$ combines the feature depending on the learnable parameter matrix $\mathbf{W}^{(l)}$.
    Building upon this, recent advances in DiGNNs further refine the message-passing scheme to capture the inherent directionality of the digraph.
    DGCN~\cite{tong2020dgcn} incorporates both first- and second-order neighbor proximity into the message aggregation process.
    DIMPA~\cite{he2022dimpa} expands the receptive field by aggregating features from $K$-hop neighborhoods.
    Inspired by the 1-WL graph isomorphism test, NSTE~\cite{kollias2022nste} tailors the message propagation to the directed nature of the graph.
    DiGCN~\cite{tong2020digcn} leverages neighbor proximity to increase the receptive field and proposes a digraph Laplacian based on personalized PageRank.
    ADPA~\cite{sun2024breaking} adaptively explores directed patterns to conduct weight-free propagation and employs two hierarchical node-wise attention mechanisms to learn representations.

\subsubsection{Spectral-based Methods}
\label{sec: Spectral-based Methods}
    Spectral-based approaches for the DiGNN depart from the strict symmetric message-passing used for undirected graphs~\cite{he2021bernnet, pmlr2022Jacobigcn}. The core of spectral-based DiGNNs is to leverage a symmetric or conjugated digraph Laplacian $\mathbf{L}_d$, which is constructed based on the directed adjacency matrix $\mathbf{A}_d$. This symmetric Laplacian $\mathbf{L}_d$ allows the application of spectral convolution operations, which can be formally represented as a function of the eigenvalues and eigenvectors of $\mathbf{L}_d$. Specifically, the layer-wise node embeddings $\mathbf{X}^{(l)}$ are computed via a first-order approximation of Chebyshev polynomials, leveraging the spectral decomposition of the symmetric digraph Laplacian.
    \begin{equation}
    \begin{aligned}
    &\;\;\;\;\;\;\;\;\mathbf{L}_d = \operatorname{DGS}(\mathbf{A}_d, \alpha, q), \\
    &\mathbf{X}^{(l+1)} = \operatorname{Poly}(\mathbf{L}_d) \operatorname{MLP}(\mathbf{X}^{(l)}),
    \end{aligned}
    \end{equation}
    where $\operatorname{DGS}(\cdot)$ is the digraph generalized symmetric function with parameters and $\operatorname{Poly}(\cdot)$ is a polynomial-based approximation method.

    Building on this, DiGCN~\cite{tong2020digcn} proposes a $\alpha$-parameterized stable state distribution based on the personalized PageRank to achieve the digraph convolution.
    MagNet~\cite{zhang2021magnet} utilizes $q$-parameterized complex Hermitian matrix to model directed information in digraphs.
    MGC~\cite{zhang2021mgc} adopts a truncated variant of PageRank, designing low-pass and high-pass filters tailored for homogeneous and heterogeneous digraphs.
    LightDiC~\cite{li2024lightdic} decouples graph propagation and feature aggregation for scalability in large-scale scenarios.

\subsection{Path-based Graph Neural Networks}
\label{sec: Path-based Graph Neural Networks}
    PathGNNs offer an effective approach to capture the intricate graph patterns by sampling and aggregating information along paths.
    For instance, GeniePath~\cite{liu2019geniepath} introduces an adaptive path layer that navigates the exploration of both the breadth and depth of the node's receptive fields.
    SPAGAN~\cite{yang2021spagan} leverages the shortest paths and applies path-based attention mechanisms to obtain node embeddings.
    PathNet~\cite{sun2022beyond} utilizes a maximal entropy-based random walk strategy to capture the heterophilous and structural information.
    RAWGNN~\cite{ijcai22_rawgnn} employs Node2Vec~\cite{Node2vec} to simulate both BFS and DFS, capturing both homophily and heterophily information.
    PathMLP~\cite{xie2023pathmlp} designs a similarity-based path sampling strategy to capture smooth paths containing high-order homophily.

\begin{figure*}[t]
\centering
\includegraphics[width=\linewidth]{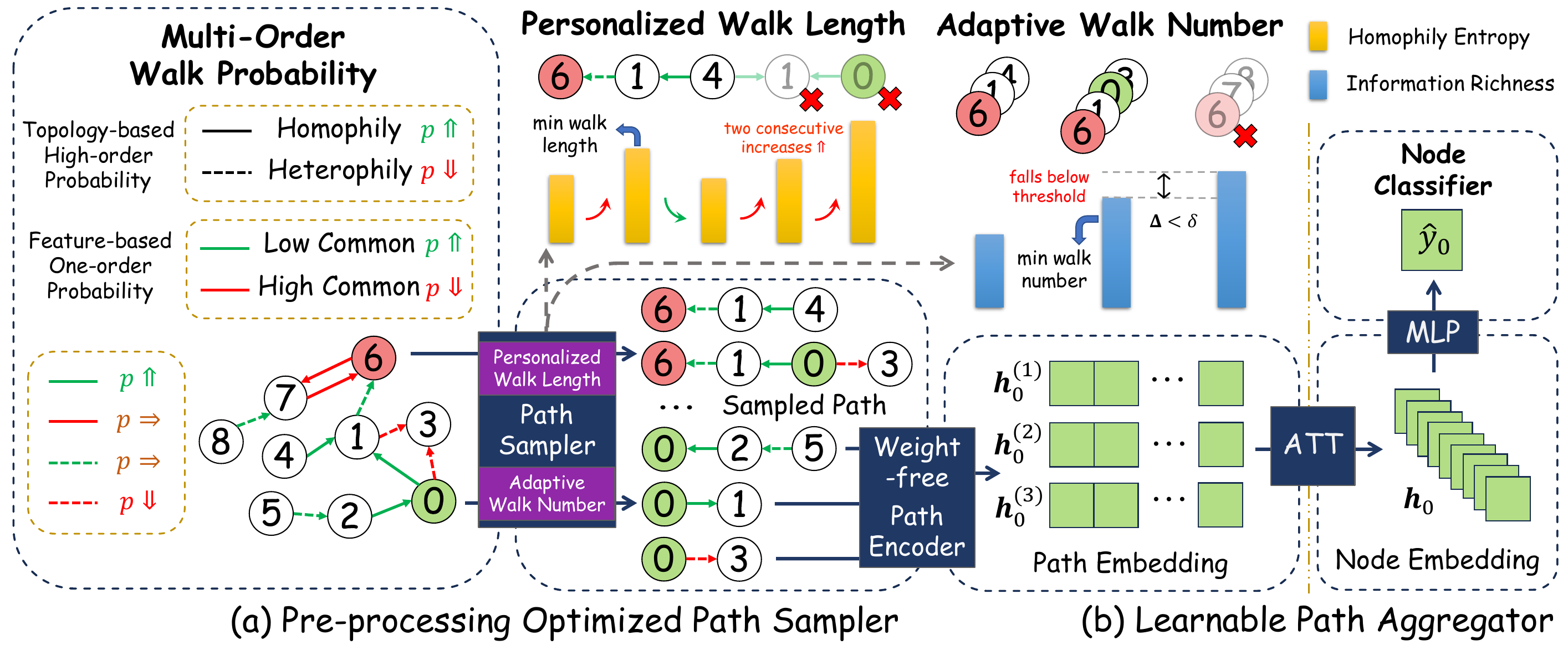}
\vspace{-5mm}
\caption{Overview of our DiRW, including (a) sampling optimized sequences and calculating the path embeddings in a weight-free manner; (b) learning the importance of different paths to obtain accurate node predictions by node-wise path aggregator.}
\label{fig: model}
\end{figure*}

\section{Method}
\label{sec: Methods}

    Building upon the key insights discussed in Sec.~\ref{sec: Introduction}, we now present our DiRW model, which is composed of two principal components: the optimized path sampler and the node-wise learnable path aggregator. The architecture of the model is depicted in Fig.~\ref{fig: model}.
    Specifically, drawing upon \textbf{Key Insight 1},  DiRW is initiated with a direction-aware path sampling strategy in Sec.~\ref{sec: Direction-aware Path Sampler}.
    Furthermore, we introduce a multi-order walk probability scheme in Sec.~\ref{sec: Multi-scale Walk Probability}.
    % , which models high-order and first-order homophily from both topological structure and node profiles.
    Guided by \textbf{Key Insight 2}, we evaluate the quality of walk sequences by introducing the homophily entropy to customize the walk length in Sec.~\ref{sec: Homophily Entropy based Personalized Walk Length}.
    In Sec.~\ref{sec: Weight-free Path Encoding}, we propose a weight-free aggregation mechanism for the path embedding.
    We also tailor walk numbers in Sec.~\ref{sec: Heuristic Walk Numbers} by assessing the information richness of sampled walks.
    In the learning phase, we integrate an attention mechanism to learn the importance of each walk sequence in Sec.~\ref{sec: Attention-based Node Encoding}.
    Culminating in Sec.~\ref{sec: Node Classifier}, we utilize a linear layer to distill the final node representations, applied to node classification and link prediction.

\subsection{Optimized Path Sampler}

\subsubsection{Direction-aware Path Sampler}
\label{sec: Direction-aware Path Sampler}
    In DiRW, edge directionality governs the walk trajectory through the direction-aware sampling protocol that dynamically adapts to node connectivity patterns.
    This critical stage is tasked with determining the destination set for each walk, which is essential for calculating the node probabilities and ensuring the continuity of the walk without interruptions.

    Specifically, when the walk encounters a sink node (node with empty in-neighborhoods), it strategically transitions to its out-neighbors. 
    Conversely, when encountering a source node (node lacking out-edges), the walk remains confined to its in-neighborhood.
    This approach effectively circumvents the issue of walk termination at nodes with unidirectional connectivity, which is mentioned in \textbf{Key Insight 1}.
    For nodes that possess both incoming and outgoing edges, a more nuanced strategy is needed to balance the edge presence and directionality.
    Drawing inspiration from the magnetic Laplacian~\cite{zhang2021magnet}, DiRW introduces a direction control parameter $q\in[0,1]$ with a uniformly sampled threshold $r\in[0,1]$ to determine the successor selection.
    If $r>q$, the walk proceeds directionally to out-neighbors, whereas $r\leq q$ permits bidirectional exploration considering both the in-neighbors and out-neighbors.
    This mechanism interpolates between strict directionality preservation ($q=0$) and undirected-like exploration ($q=1$), while preventing walk termination at unidirectionally connected nodes.

\subsubsection{Multi-order Walk Probability}
\label{sec: Multi-scale Walk Probability}
    The computation of walk probabilities in DiRW is a nuanced process that seamlessly integrates both the topological structure and node profiles, enabling DiRW to delve into the complexities of both high-order and one-order homophily within heterophilous graphs.

    \textbf{Topology-based High-order Probability.}
    DiRW uncovers high-order homophily information which is often embedded within heterophilous graphs~\cite{xie2023pathmlp} by strategically prioritizing nodes with few common neighbors, thus bridging to distant areas of the graph.
    The topology-based high-order probability $\mathbf{P}^{topo}_u$ is inversely proportional to the number of common neighbors $\text{Com}(u,v)$ and normalized by the degree $\text{Deg}(v)$ of the candidate node $v$:
    \begin{equation}
    \mathbf{P}^{topo}_u(v)=1-\frac{\text{Com}\left(u,v\right)+1}{\text{Deg}(v)},\;\;\text{if}\;(u,v)\in\mathcal{E}.
    \end{equation}

    \textbf{Feature-based One-order Probability.}
    From the feature standpoint, DiRW captures one-order homophily by favoring walks towards neighbors with more similar features, which indicate stronger homophily.
    The feature-based one-order probability $\mathbf{P}^{feat}_u$ is calculated based on the cosine similarity between the features of the current node $u$ and candidate node $v$:
    \begin{equation}
    \mathbf{P}^{feat}_u(v)=\cos\left(\mathbf{X}_{u},\mathbf{X}_{v}\right),\;\;\text{if}\;(u,v)\in\mathcal{E}.
    \end{equation}

    The transition probability from node $u$ to node $v$ in a walk step is determined by combining the topology-based and feature-based probabilities, normalized by a softmax function:
    \begin{equation}
    \mathbf{P}^{dst}_u(v)={\rm Softmax}\left(\mathbf{P}^{topo}_u\left(v\right)+\mathbf{P}^{feat}_u\left(v\right)\right).
    \end{equation}
    Recognizing that the softmax scales vary with the walk directions, and we pre-compute $\mathbf{P}^{topo}$ and $\mathbf{P}^{feat}$ to ensure efficiency.

\subsubsection{Homophily Entropy-based Personalized Walk Length}
\label{sec: Homophily Entropy based Personalized Walk Length}
    In DiRW, the walk length is dynamically determined based on homophily entropy rather than being predefined and fixed.
    Traditional homophily metrics predominantly focus on immediate neighborhood information, providing a limited view of the walk sequence's quality.
    To address this limitation, we introduce a novel metric \textit{Homophily Entropy}, which assesses the walk sequence quality based on feature similarity among nodes in the sequence.
    Our objective is to ensure sampled nodes in the walk sequence have a strong homophilous relationship with the originating node, enriching the aggregated information with relevance to it.
    We first convert the walk sequence into a homophily label sequence, represented as:    
    \begin{equation}
    S^{homo}(i) = \mathbb{I} \left(\mathbf{X}_{S^{walk}(i)}\neq\mathbf{X}_{S^{walk}(0)}\right) \cdot i,
    \end{equation}
    where $\mathbb{I}(\text{condition})$ equals $1$ if the condition is met and $0$ otherwise.
    $S^{walk}(i)$ denotes the $i$-th node in the walk sequence, and $S^{homo}(i)$ is the corresponding homophily label.
    By applying the Shannon entropy~\cite{li2016structural_entropy_toit_16} on $S^{homo}$, we derive the homophily entropy:
    \begin{equation}
    \label{eq: Homophily Entropy}
    H_{homo}\left(S^{walk}\right)=H\left(S^{homo}\right)=-\sum_{i=1}^k p(i)\log p(i),
    \end{equation}
    where $p(i)$ represents the probability of encountering $S^{homo}(i)$ in $S^{homo}$ and $k$ denotes the length of the sequence. 
    Our analysis indicates that a lower homophily entropy corresponds to a higher walk sequence quality.
    
    This entropy metric serves for dynamically determining node-specific walk lengths.
    We start with a predefined minimum walk length $l_{min}$ to guarantee the adequate information acquisition.
    Subsequent walk persists until two consecutive homophily entropy increments, signaling the heterophilous neighborhood.
    This adaptive scheme aligns with \textbf{Key Insight 2}, which explicitly accommodates nodes with homophilous neighborhoods.

\begin{table*}[t]
\caption{The statistician of the experimental datasets.}
\vspace{-3mm}
\label{tab: datasets}
\resizebox{\textwidth}{!}{
\begin{tabular}{c|c|cccccc}
\toprule
Characteristics & Datasets & \#Nodes & \#Edges & \#Features & \#Node Classes & \#Train/Val/Test & \#Description \\ 
\midrule
\multirow{5}{*}{Directed-Homophily} & CoraML & 2,995 & 8,416 & 2,879 & 7 & 140/500/2,355 & citation network\\
& CiteSeer & 3,312 & 4,591 & 3,703 & 6 & 120/500/2,692 & citation network\\
& WikiCS & 11,701 & 290,519 & 300 & 10 & 580/1,769/5,847 & weblink network\\
& Amazon-Computers & 13,752 & 287,209 & 767 & 10 & 200/300/12,881 & co-purchase network\\
& ogbn-arxiv & 169,343 & 2,315,598 & 128 & 40 & 91k/30k/48k & citation network\\ 
\midrule
UnDirected-Homophily & ogbn-products & 2,449,029 &  61,859,140 & 100 & 47 & 196k/49k/2204k &  co-purchase network\\ 
\midrule
\multirow{3}{*}{Directed-Heterophily} & Chameleon & 890 & 13,584 & 2,325 & 5 & 48\%/32\%/20\% & wiki-page network\\
& Actor & 7,600 & 26,659 & 932 & 5 & 48\%/32\%/20\% & actor network\\
& Rating & 24,492 & 93,050 & 300 & 5 & 50\%/25\%/25\% & rating network\\ 
\bottomrule
\end{tabular}}
\vspace{-3mm}
\end{table*}

\subsubsection{Weight-free Path Encoding}
\label{sec: Weight-free Path Encoding}
    To address the varying influence of nodes along a path, DiRW uses an exponential decay function to assign weights and compute the path embedding:
    \begin{equation}
    \mathbf{h}_u^{(l)}=\sum_{i=1}^k \frac{\gamma^i}{\sum_{j=1}^k\gamma^{j}}\mathbf{X}_{P_u^{(l)}(i)},
    \end{equation}
    where $\gamma\in (0,1)$ represents the decay parameter and $P_u^{(l)}$ denotes the $l$-th sampled path of node $u$.
    This weight-free aggregation mechanism allows for the integration of the sampling into the pre-processing stage, which substantially reduces the computational complexity during training.
    Moreover, by giving greater significance to nodes closer to the originating node, the exponential decay mechanism effectively captures the sequential information inherent in the walk sequence, overcoming the constraints of uniform weighting often found in message-passing approaches~\cite{kipf2016gcn}.

\subsubsection{Adaptive Walk Numbers}
\label{sec: Heuristic Walk Numbers}
    In DiRW, gathering comprehensive context is essential, which influences the walk numbers per node.
    This is determined by assessing the \textit{Information Richness}, measured by the average of sampled path embeddings.
    To avoid early walk termination, a minimum walk number $n_{min}$ is set.
    After each walk, DiRW computes the L2 norm of the Information Richness difference between the current and previous sampled sequences:
    \begin{equation}
    \Delta=\left|\left|\frac{1}{L}\sum_{l=1}^L\mathbf{h}_u^{(l)} - \frac{1}{L-1}\sum_{l=1}^{L-1} \mathbf{h}_u^{(l)}\right|\right|_2,\;\; L>n_{min},
    \end{equation}
    where $L$ is the number of sampled path sequences.
    
    If $\Delta$ falls below the predefined threshold $\delta$, it indicates that the sampling has reached a saturation point, beyond which additional walks are unlikely to contribute substantially to the information richness.
    In such cases, the walk is stopped to avoid further unnecessary and redundant sampling, improving the information gathering efficiency and reflecting \textbf{Key Insight 2}: heterophilous nodes are often characterized by more intricate neighborhoods and necessitate more walks to amass sufficiently rich information.

\subsection{Learnable Path Aggregator}

\subsubsection{Attention-based Node Encoding}
\label{sec: Attention-based Node Encoding}
    After the pre-processing sampling phase, DiRW harnesses an attention mechanism to discern and weigh the informativeness of various path embeddings.
    This process is pivotal for constructing node embeddings that are rich in relevant contextual information.

    Specifically, for each node $u$, DiRW initiates the encoding of the sampled paths $\mathbf{h}_{u}$ 
    through a pair of linear layers, interspersed with a LeakyReLU activation function.
    The encoded representation $\mathbf{e}_u$ is subsequently processed through a softmax function to yield attention scores $\alpha_u$.
    The final node embedding $\mathbf{z}_u$ is then computed as a weighted aggregation of the path embeddings $\mathbf{h}_u^{(i)}$, with the weights being the attention scores $\alpha_u^i$, articulated as follows:
    \begin{equation}
    \begin{aligned}
    \label{eq: Attention}
    &\mathbf{e}_u={\rm MLP}_2\left({\rm LeakyReLu}\left({\rm MLP}_1\left(\mathbf{h}_{u}\right)\right)\right), \\
    &\;\alpha_u={\rm Softmax}\left(\mathbf{e}_u\right), \;\;\mathbf{z}_u=\sum_{l=1}^L\alpha_u^l\mathbf{h}_u^{(l)}.
    \end{aligned}
    \end{equation}

\subsubsection{Node Classifier}
\label{sec: Node Classifier}
    With the node embeddings $\mathbf{z}_u$ at hand, DiRW deploys an MLP to tackle the node classification task.
    The training process is directed by the cross-entropy loss as follows:
    \begin{equation}
    \label{eq: classifier}
    \hat{\mathbf{y}}_u={\rm Softmax}\left({\rm MLP}_3\left(\mathbf{z}_u\right)\right),\;\;\mathcal{L}=-\frac{1}{|\mathcal{V}_l|}\sum_{i\in\mathcal{V}_l}\mathbf{Y}_i\log\hat{\mathbf{y}}_i,
    \end{equation}
    where $\mathcal{V}_l$ represents the training set. 
    $\mathbf{Y}_i$ and $\hat{\mathbf{y}}_i$ are the one-hot encoded true label and the predicted label of node $i$, respectively.

    In the link prediction task, DiRW estimates the probability of edges between node pairs using another MLP.
    The input to this MLP consists of the concatenated embeddings of the node pairs, allowing the model to leverage the feature information from both nodes.
    The training process is also guided by the binary cross-entropy loss.

\section{Experiments}
\label{sec: Experiments}

    In this section, we conduct a comprehensive evaluation of our DiRW, structured to address five  critical research dimensions:
    \textbf{Q1}:
    Can DiRW establish superior performance as both a new DiGNN and a plug-and-play strategy?
    \textbf{Q2}:
    If DiRW proves effective, what architectural characteristics contribute to its enhanced performance?
    \textbf{Q3}:
    How does DiRW's time complexity compare with baseline methods in practical deployment scenarios?
    \textbf{Q4}:
    How does the hyperparameters sensitivity and robustness of DiRW manifest?
    \textbf{Q5}:
    How does a comparative analysis of path quality between DiRW and existing PathGNNs inform its structural learning advantages?
    
\subsection{Experimental Setup}

\subsubsection{Datasets.}

    Our comprehensive evaluation spans 9 benchmark datasets encompassing both directed and undirected graphs with diverse homophily characteristics.
    We have documented the detailed information of these datasets in the Tab.~\ref{tab: datasets}.

\subsubsection{Baselines.}

    Our experiment leverages diverse GNNs as comparative benchmarks, which can be categorized as follows:
    (i) traditional undirected approaches: GCN~\cite{kipf2016gcn}, GAT~\cite{velivckovic2017gat};
    (ii) directed spatial methods: DGCN~\cite{tong2020dgcn}, NSTE~\cite{kollias2022nste}, DIMPA~\cite{he2022dimpa}, Dir-GNN~\cite{dirgnn_rossi_2023}, ADPA~\cite{sun2024breaking};
    (iii) directed spectral methods: DiGCN~\cite{tong2020digcn} and its variant DiGCNappr, MagNet~\cite{zhang2021magnet}, MGC~\cite{zhang2021mgc}, LightDiC~\cite{li2024lightdic};
    (iiii) PathGNNs: RAWGNN~\cite{ijcai22_rawgnn} and PathNet~\cite{sun2022beyond}.
    To minimize randomness and ensure fair comparisons, we repeated each experiment 10 times to obtain unbiased performances.
    Moreover, we transform the digraph to undirected graph and feed it into undirected GNNs.

\subsubsection{Hyperparameter Settings.}

    The hyperparameters in the baseline GNNs are set following the original paper if available.
    Otherwise, we perform a hyperparameter search via the Optuna~\cite{akiba2019optuna}.
    In DiRW, we perform a grid search for the minimum walking length ranging from $2$ to $6$, and for the minimum walk number ranging from $2$ to $10$. 
    Furthermore, we fine-tune the walk direction coefficient $q$ within the interval $[0.5, 1]$.

\subsubsection{Experimental Environment.}

    Our experiments are conducted on the machine with Intel(R) Xeon(R) Platinum 8468V, NVIDIA H800 PCIe, and CUDA 12.2. 
    The operating system is Ubuntu 20.04.6.
    As for the software, we use Python 3.8 and Pytorch 2.2.1.

    \begin{table*}[t]
    \caption{Model performance (\%) as a new DiGNN in node classification. The best result is \colorbox{blue!15!white}{\textbf{bold}}. The second result is \underline{underlined}.}
    \vspace{-2.5mm}
    \label{tab: dirw overall performance}
    \resizebox{\textwidth}{!}{
    \begin{tabular}{cc|cccccccccc}
    \toprule
    \multirow{2}{*}{Type} & \multirow{2}{*}{Models} & \multirow{2}{*}{CoraML} & \multirow{2}{*}{CiteSeer} & \multirow{2}{*}{WikiCS} & Amazon & \multirow{2}{*}{Chameleon} & \multirow{2}{*}{Actor} & \multirow{2}{*}{Rating} & \multirow{2}{*}{Arxiv} & \multirow{2}{*}{Products} & \multirow{2}{*}{Rank} \\
    &&&&& Computers &&&&&&\\ 
    \midrule
    Undirected & GCN & \colorbox{blue!15!white}{\textbf{84.48$\pm$0.17}} & 65.26$\pm$1.07 & 78.98$\pm$0.49 & 83.26$\pm$0.51 & 40.93$\pm$1.24 & 30.96$\pm$0.45 & 43.37$\pm$0.13 & 67.80$\pm$0.07 & 73.3$\pm$0.07 & 5.67 \\
    GNNs & GAT & 83.73$\pm$0.47 & 63.12$\pm$1.06 & 79.35$\pm$0.28 & 80.16$\pm$2.32 & 40.82$\pm$2.46 & 30.29$\pm$0.37 & 45.03$\pm$0.50 & \underline{67.79$\pm$0.24} & OOM & 7.44 \\ 
    \midrule
    & DGCN & 81.27$\pm$0.58 & 64.66$\pm$0.65 & 78.71$\pm$0.19 & 82.95$\pm$1.13 & 41.96$\pm$1.00 & 30.87$\pm$0.38 & 44.24$\pm$0.28 & OOT & OOM & 8.00 \\
    \multirow{2}{*}{Spatial} & NSTE & 80.06$\pm$0.89 & 62.99$\pm$1.08 & 77.58$\pm$0.29 & 78.50$\pm$1.87 & 40.31$\pm$1.87 & 31.14$\pm$0.46 & 45.31$\pm$0.15 & 64.40$\pm$0.94 & OOT & 9.78 \\
    \multirow{2}{*}{DiGNNs} & DIMPA & 79.77$\pm$0.66 & 60.88$\pm$1.08 & 78.58$\pm$0.27 & 77.80$\pm$1.42 & 40.41$\pm$1.88 & 30.84$\pm$0.73 & 41.58$\pm$0.17 & 65.89$\pm$0.22 & 72.89$\pm$0.13 & 10.67 \\ 
    & Dir-GNN & 79.92$\pm$1.26 & 63.74$\pm$0.87 & 79.04$\pm$0.26 & 81.34$\pm$0.92 & 41.96$\pm$1.95 & 30.12$\pm$0.65 & 43.74$\pm$0.10 & 64.26$\pm$0.48 & 72.93$\pm$0.17 & 8.33 \\
    & ADPA & 79.10$\pm$1.56 & 65.01$\pm$2.29 & \underline{79.43$\pm$0.92} & 73.25$\pm$4.26 & 41.66$\pm$3.07 & 35.29$\pm$0.82 & 44.90$\pm$0.62 & 64.28$\pm$0.56 & OOT & 7.67 \\ 
    \midrule
    & DiGCN & 80.81$\pm$0.52 & 62.38$\pm$0.28 & 79.30$\pm$0.32 & 80.02$\pm$1.52 & 41.24$\pm$1.67 & 34.30$\pm$0.89 & \underline{45.99$\pm$0.28} & 65.38$\pm$0.47 & OOM & 7.11 \\ 
    \multirow{2}{*}{Spectral} & DiGCNappr & 80.65$\pm$0.40 & 62.05$\pm$0.59 & 78.54$\pm$0.32 & 81.53$\pm$1.27 & 40.93$\pm$2.49 & 30.99$\pm$0.49 & 40.46$\pm$0.20 & 61.57$\pm$0.56 & 73.51$\pm$0.05 & 9.33 \\
    \multirow{2}{*}{DiGNNs} & MagNet & 80.45$\pm$0.46 & \underline{65.63$\pm$1.69} & 79.19$\pm$0.20 & \underline{83.41$\pm$0.99} & \underline{42.78$\pm$0.63} & 31.39$\pm$0.53 & 45.97$\pm$0.71 & 68.22$\pm$0.25 & \underline{74.21$\pm$0.06} & 3.78 \\
    & MGC & 82.66$\pm$1.01 & 64.78$\pm$1.05 & 77.42$\pm$0.28 & 81.26$\pm$0.80 & 41.34$\pm$3.29 & 35.75$\pm$0.41 & 42.01$\pm$0.36 & 66.73$\pm$0.17 & 59.86$\pm$3.20 & 7.11 \\ 
    & LightDiC & 82.67$\pm$1.10 & 64.36$\pm$0.35 & 78.02$\pm$0.21 & 79.87$\pm$0.68 & 42.68$\pm$1.61 & 30.59$\pm$0.50 & 40.53$\pm$0.48 & 63.41$\pm$0.26 & 64.85$\pm$0.07 & 8.78 \\
    \midrule
    & RAWGNN & 78.35$\pm$0.90 & 59.94$\pm$0.38 & 76.76$\pm$0.31 & 72.76$\pm$0.67 &  41.24$\pm$3.34 & 34.71$\pm$0.80 & 44.07$\pm$0.32 & OOM & OOM & 11.89 \\
    PathGNNs & PathNet & 66.61$\pm$6.83 & 61.41$\pm$1.93 & 75.71$\pm$0.66 & 68.78$\pm$2.13 & 38.44$\pm$3.21 & \colorbox{blue!15!white}{\textbf{36.67$\pm$0.92}} & 39.15$\pm$0.40 & OOM & OOM & 13.22 \\
    & \textbf{DiRW} & \underline{84.24$\pm$0.19} & \colorbox{blue!15!white}{\textbf{65.88$\pm$1.41}} & \colorbox{blue!15!white}{\textbf{79.87$\pm$0.08}} & \colorbox{blue!15!white}{\textbf{83.50$\pm$0.68}} & \colorbox{blue!15!white}{\textbf{43.92$\pm$0.23}} & \underline{36.42$\pm$0.46} & \colorbox{blue!15!white}{\textbf{46.13$\pm$0.39}} & \colorbox{blue!15!white}{\textbf{68.51$\pm$0.18}} & \colorbox{blue!15!white}{\textbf{75.19$\pm$0.47}} & 1.22 \\ 
    \bottomrule
    \end{tabular}}
    \end{table*}

    \begin{table*}[t]
    \caption{Model Performance (\%) as a plug-and-play strategy on PathGNNs in node classification.}
    \vspace{-2.5mm}
    \label{tab: pathgnns}
    \resizebox{\textwidth}{!}{
    \begin{tabular}{c|ccccccccc}
    \toprule
    \multirow{2}{*}{Models} & \multirow{2}{*}{CoraML} & \multirow{2}{*}{CiteSeer} & \multirow{2}{*}{WikiCS} & Amazon & \multirow{2}{*}{Chameleon} & \multirow{2}{*}{Actor} & \multirow{2}{*}{Rating} & \multirow{2}{*}{Improvement} \\
    &&&& Computers &&&&\\ 
    \midrule
    RAWGNN & 78.35$\pm$0.90 & 59.94$\pm$0.38 & 76.76$\pm$0.31 & 72.76$\pm$0.67 &41.24$\pm$3.34 & 34.71$\pm$0.80 & 44.07$\pm$0.32 & \multirow{2}{*}{\textcolor{red}{$\Uparrow 4.34\%$}}\\
    RAWGNN+DiRW & 82.67$\pm$0.26 & 64.54$\pm$0.75 &  77.71$\pm$0.27 & 76.33$\pm$1.02 & 43.30$\pm$1.31 & 35.55$\pm$0.39 & 45.66$\pm$0.63 & \\ 
    \midrule
    PathNet & 66.61$\pm$6.83 & 61.41$\pm$1.93 & 75.71$\pm$0.66 & 68.78$\pm$2.13 & 38.44$\pm$3.21 & 36.67$\pm$0.92 & 39.15$\pm$0.40 & \multirow{2}{*}{\textcolor{red}{$\Uparrow 1.25\%$}}\\
    PathNet+DiRW & 68.25$\pm$0.77 & 62.66$\pm$0.89 & 76.21$\pm$0.35 & 69.45$\pm$0.97 & 38.88$\pm$1.04 & 37.14$\pm$0.87 & 39.22$\pm$0.24 & \\ 
    \midrule
    DiRW & \textbf{84.24$\pm$0.19} & \textbf{65.88$\pm$1.41} & \textbf{79.87$\pm$0.08} & \textbf{83.50$\pm$0.68} & \textbf{43.92$\pm$0.23} & \textbf{36.42$\pm$0.46} & \textbf{46.13$\pm$0.39} & --- \\ 
    \bottomrule
    \end{tabular}}
    \vspace{-2mm}
    \end{table*}

\subsection{Overall Performance}

\subsubsection{An Innovative Learning Architecture}

    To answer \textbf{Q1}, we conduct comparative experiments to evaluate the performance of DiRW in node classification.
    The results presented in Tab.~\ref{tab: dirw overall performance} demonstrate that, as a novel DiGNN, DiRW achieves exceptional performance across all datasets, outperforming the leading DiGNN, MagNet, by an average of 3.2\%.
    This improvement is attributed to its effective modeling of digraphs and heterophilous relationships.

    In contrast, traditional undirected GNNs perform well in homophilous scenarios but struggle in heterophilous digraphs due to their reliance on simplistic undirected adjacency matrix approaches.
    While Dir-GNN and ADPA attempt to address heterophily through digraph modeling, their methods exhibit limitations in generalizing effectively to homophilous digraphs.
    Furthermore, although undirected PathGNNs show potential in modeling heterophilous graphs, their performance is significantly compromised by neglect of edge direction and simplistic walk strategies.
    Notably, they fail to account for edge directionality in digraphs, and their simplistic walk strategies limit their effectiveness in complex digraph structures.

    Emphasize that the sophisticated model architectures like RNNs in PathGNNs cause scalability issues, leading to out-of-memory (OOM) errors with large-scale graphs like ogbn-arxiv and ogbn-products. 
    Similarly, the intricate message-passing paradigm with multiple convolutional layers in DGCN and NSTE often leads to incomplete training within 12 hours, resulting in out-of-time (OOT) errors.
    In contrast, DiRW demonstrates high efficiency and superior performance on large-scale digraphs through its weight-free path sampling strategy and lightweight learning mechanism.

\subsubsection{A Plug-and-Play Approach}
    In addition to evaluating DiRW as an innovative neural architecture, we also integrate its optimized path sampler with other PathGNNs, specifically combining DiRW with RAWGNN and PathNet for digraph modeling.
    However, due to the complicated path and node embedding learning rules designed in these two models, it is tough for the weight-free adaptive walking length and number in DiRW to generalize to them.
    Therefore, we only incorporated the direction-aware sampling strategy and walking probabilities that consider both topological structure and node profiles into these two models to assess their performance.
    The results are demonstrated in Tab.~\ref{tab: pathgnns}.

    The results provide strong evidence that our optimized path sampler significantly enhances the performance of both RAWGNN and PathNet.
    The performance gains stem from the shortcomings of the original PathGNNs, which fail to consider adequately the directionality of edges in digraphs, leading to a substantial loss of valuable information.
    DiRW corrects it by balancing the edge directionality and existence into its sampling process, effectively capturing the intricate topological structures inherent in digraphs.
    Moreover, our walking probabilities, which account for topological structure and node profiles, outperform those of RAWGNN and PathNet, which focus solely on topology.
    By modeling node similarity, DiRW effectively captures one-hop homophily in heterophilous contexts.
    This integration not only enhances performance when DiRW is used as a plug-and-play module with other PathGNNs but also highlights the adaptability and robustness of our approach.

\subsubsection{Performance in Link Prediction}
    
    \begin{table*}[t]
    \caption{Model performance (\%) in link direction prediction. The best result is \colorbox{blue!15!white}{\textbf{bold}}. The second result is \underline{underlined}.}
    \vspace{-2.5mm}
    \label{tab: dirw link performance}
    \resizebox{\textwidth}{!}{
    \begin{tabular}{c|cccccccc}
    \toprule
    \multirow{2}{*}{Models} & \multirow{2}{*}{CoraML} & \multirow{2}{*}{CiteSeer} & \multirow{2}{*}{WikiCS} & Amazon & \multirow{2}{*}{Chameleon} & \multirow{2}{*}{Actor} & \multirow{2}{*}{Rating} & \multirow{2}{*}{Arxiv} \\
    &&&& Computers &&&&\\ 
    \midrule
    GCN & 83.63$\pm$0.74 & 74.29$\pm$0.73 & 72.97$\pm$0.23 & 92.42$\pm$0.14 & 87.98$\pm$0.48 & 58.37$\pm$1.34 & 50.16$\pm$0.74 & 82.24$\pm$0.12 \\
    GAT & 81.28$\pm$3.40 & 79.33$\pm$2.24 & 70.18$\pm$9.71 & 67.35$\pm$0.85 & 85.80$\pm$1.83 & 54.13$\pm$0.42 & 50.84$\pm$0.88 & 76.55$\pm$4.46 \\
    DGCN & 89.47$\pm$0.25 & 85.09$\pm$0.29 & 87.02$\pm$0.08 & 95.67$\pm$0.13 & 91.38$\pm$0.58 & 82.58$\pm$0.66 & 78.96$\pm$0.10 & OOT \\
    DiGCN & 88.15$\pm$0.57 & 87.99$\pm$1.77 & 85.38$\pm$0.19 & 96.45$\pm$0.09 & 89.84$\pm$0.61 & 82.77$\pm$0.44 & 80.92$\pm$0.05 & 92.71$\pm$0.04 \\ 
    DiGCNappr & 84.96$\pm$0.23 & 82.50$\pm$0.51 & 83.41$\pm$0.10 & 95.72$\pm$0.07 & 91.54$\pm$1.25 & 81.60$\pm$0.25 & 78.92$\pm$0.06 & 91.71$\pm$0.01 \\
    NSTE & 90.10$\pm$0.59 & 87.86$\pm$0.60 & 87.69$\pm$0.15 & 95.49$\pm$0.56 & 91.28$\pm$0.69 & 79.74$\pm$0.20 & 80.29$\pm$0.39 & 92.31$\pm$0.07 \\
    DIMPA & \underline{90.48$\pm$0.33} & 86.83$\pm$0.50 & 82.54$\pm$0.15 & 95.65$\pm$0.16 & 89.68$\pm$0.69 & 82.33$\pm$0.24 & 83.06$\pm$0.06 & 93.56$\pm$0.12 \\
    MagNet & 89.25$\pm$0.34 & 85.13$\pm$0.60 & 89.07$\pm$0.10 & 96.73$\pm$0.07 & 90.64$\pm$1.18 & 83.69$\pm$0.39 & 82.75$\pm$0.18 & 93.91$\pm$0.04 \\
    MGC & 89.97$\pm$0.64 & 81.79$\pm$1.63 & 86.66$\pm$0.17 & 86.86$\pm$0.46 & 89.15$\pm$0.97 & 67.84$\pm$0.37 & 72.39$\pm$0.17 & 87.65$\pm$4.27 \\ 
    LightDiC & 88.87$\pm$0.46 & 87.99$\pm$0.46 & 88.77$\pm$0.06 & 95.30$\pm$0.05 & 90.27$\pm$0.36 & 82.39$\pm$0.35 & 81.59$\pm$0.07 & 92.73$\pm$0.01 \\
    Dir-GNN & 88.95$\pm$0.45 & 86.34$\pm$0.99 & \colorbox{blue!15!white}{\textbf{88.87$\pm$0.05}} & \underline{96.93$\pm$0.03} & \underline{91.97$\pm$0.63} & \colorbox{blue!15!white}{\textbf{85.89$\pm$0.17}} & \underline{85.01$\pm$0.05} & \underline{94.73$\pm$0.05} \\
    ADPA & 90.28$\pm$0.78 & \underline{88.57$\pm$0.68} & \underline{88.85$\pm$0.14} & 96.66$\pm$0.06 & 89.52$\pm$0.83 & 84.88$\pm$0.34 & 82.69$\pm$0.24 & 93.81$\pm$0.04\\ 
    RAWGNN & 89.92$\pm$1.21 & 87.01$\pm$1.04 & 83.48$\pm$0.8 & 94.93$\pm$0.18 & 89.95$\pm$1.21 & 85.34$\pm$0.46 & 73.26$\pm$0.43 & OOM \\ 
    \midrule
    \textbf{DiRW} & \colorbox{blue!15!white}{\textbf{91.05$\pm$0.57}} & \colorbox{blue!15!white}{\textbf{88.96$\pm$1.38}} & 88.79$\pm$0.49 & \colorbox{blue!15!white}{\textbf{97.15$\pm$0.24}} & \colorbox{blue!15!white}{\textbf{92.19$\pm$1.17}} & \underline{85.56$\pm$0.39} & \colorbox{blue!15!white}{\textbf{85.49$\pm$0.34}} & \colorbox{blue!15!white}{\textbf{95.27$\pm$0.08}} \\ 
    \bottomrule
    \end{tabular}}
    \end{table*}
    
    We extend the evaluation of DiRW through a multi-task framework, investigating both node classification and directional link prediction capabilities, with the results presented in Tab.~\ref{tab: dirw link performance}.
    We directly leverage the node embeddings derived from Eq.~(\ref{eq: Attention}) and concatenate them to form discriminative edge embeddings, thereby bypassing the need for additional node classifier as specified in Eq.~(\ref{eq: classifier}).
    This concatenated vector serves as a comprehensive representation of the edges, preserving the geometric relationships encoded during the walk aggregation phase while maintaining architectural simplicity.
    
    The experimental outcomes provide compelling evidence that DiRW achieves an average performance gain of $0.82\%$ over the second leading model Dir-GNN.
    The efficacy of DiRW in link prediction is a testament to its ability to capture the subtleties of graph structure and the nuanced relationships between nodes.
    This capability stems from the model's sophisticated path sampling strategies and attention-based aggregation mechanism, which allow for a deep understanding of the digraph's topology and node features.
    
    It is important to note that traditional undirected GNNs have exhibited subpar performance and suffer from structural isomorphism issues in direction-sensitive tasks. 
    The primary reason for this deficit lies in the direction-agnostic message-passing in undirected architectures, which fails to discriminate between ($u\rightarrow v$) and ($v\rightarrow u$) edge semantics.
    In contrast, DiRW's design explicitly accounts for edge directionality, providing it with a distinct advantage in tasks where the direction of relationships is pivotal. 
    This comparative analysis underscores the importance of adopting models that can accommodate the directed nature of graphs, thereby offering a more accurate representation of the underlying structure and relationships within the data.

\subsection{Ablation Study}

\begin{table}[t]
\caption{Ablation study performance (\%).}
\vspace{-2.5mm}
\label{tab: ablation}
\resizebox{0.48\textwidth}{!}{
\begin{tabular}{c|cccc}
\toprule
Models & CiteSeer & WikiCS & Chameleon & Rating \\ 
\midrule
w/o Dir & 65.11$\pm$1.77 & 79.65$\pm$0.18 & 42.47$\pm$1.07 & 45.59$\pm$0.16 \\
w/o Topo & 64.26$\pm$0.17 & 79.37$\pm$0.13 & 38.97$\pm$2.64 & 45.79$\pm$0.18 \\
w/o Feat & 65.77$\pm$1.28 & 79.27$\pm$0.13 & 41.65$\pm$1.73 & 45.62$\pm$0.25 \\
\midrule
w/o Att & 65.02$\pm$0.75 & 79.26$\pm$0.32 & 39.59$\pm$2.23 & 42.55$\pm$0.39 \\
DiRW-Gate & 65.65$\pm$0.84 & 79.52$\pm$0.28 & 38.76$\pm$3.56 & 42.68$\pm$0.15 \\
DiRW-JK & 63.88$\pm$1.02 & 79.31$\pm$0.25 & 40.21$\pm$3.82 & 42.74$\pm$0.33 \\
\midrule
DiRW & \textbf{65.88$\pm$1.41} & \textbf{79.87$\pm$0.08} & \textbf{43.92$\pm$0.23} & \textbf{46.13$\pm$0.39} \\ 
\bottomrule
\end{tabular}}
\end{table}
    
\begin{figure*}[t]
\centering
\includegraphics[width=\linewidth]{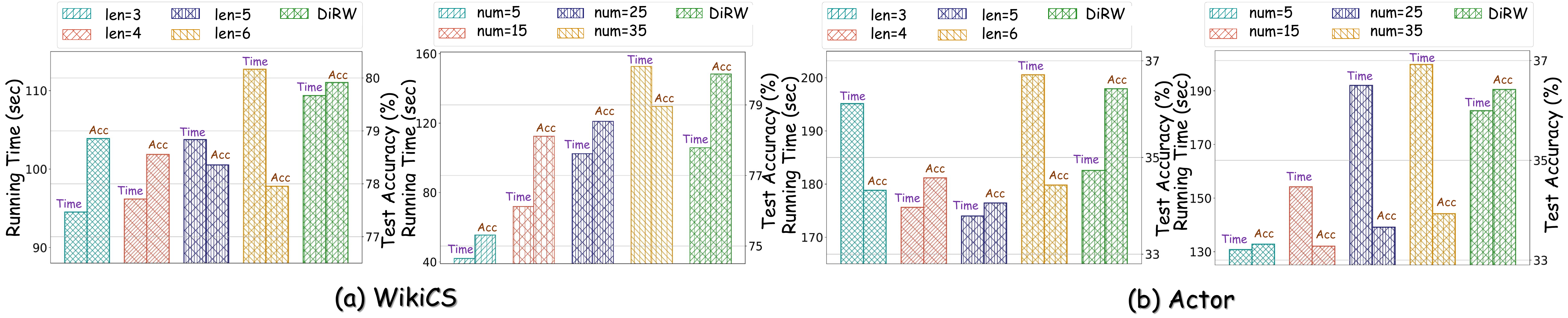}
\vspace{-6mm}
\caption{Performance and running time of DiRW and its variants w/o personalized walk length and adaptive walk number.}
\vspace{-1mm}
\label{fig: ablation}
\end{figure*}

    To answer \textbf{Q2}, we have conducted a series of ablation studies to ascertain the individual contributions of the five core components that constitute the backbone of DiRW.

    We began with the direction-aware path sampler by setting the walk direction coefficient $q$ to $1$, simulating a scenario without directionality (w/o Dir).
    Then, we conducted separate ablations for the topology-based (w/o Topo) and feature-based (w/o Feat) walk probabilities to assess the relative importance of node profiles and topological structure in guiding the walk.
    Additionally, we conducted an ablation experiment on our node-wise attention mechanisms, replacing them with direct averaging of path embeddings (w/o Att), Gate attention~\cite{ahmad2021gate} (DiRW-Gate), and JK attention~\cite{xu2018jknet} (DiRW-JK).
    The outcomes are detailed in Tab.~\ref{tab: ablation}.

    Furthermore, to assess the adaptive walk length and numbers, we replace them with predefined and fixed hyperparameters and depict the outcomes in Fig.~\ref{fig: ablation}.
    We measured the performance and running time across varying walk lengths and numbers while keeping other hyperparameters consistent with DiRW.
    From these results, we draw the following conclusions:

\textbf{Direction-aware path sampler.}
    By effectively balancing the direction and existence of edges, the direction-aware path sampler has significantly enhanced the model's capacity to capture the intricate topological structures of digraphs, leading to an average performance improvement of 1.5\%.

\textbf{Multi-order walk probability.}
    The topology-based high-order walk probability shows a clear advantage in detecting homophilous patterns, leading to significant improvements in predictive performance.
    Moreover, the feature-based one-order walk probability achieves an average performance increase of 1.9\%, underscoring feature information's importance in guiding walk preferences.

\textbf{Node-wise learnable path aggregator.}
    By utilizing our attention mechanism, DiRW effectively distinguishes between productive and unproductive sampling sequences, leading to high-quality node embeddings and an average performance boost of 5.36\%.
    The node-wise attention mechanism employs a dual MLP, which captures complex nuances in path embeddings more effectively than the single MLP used in the JK and Gate attention mechanisms.

\textbf{Homophily entropy-based personalized walk length.}
    The personalized walk length greatly enhances DiRW's performance, outperforming static walk length models that require significant pre-processing and often yield sub-optimal results.
    In contrast, DiRW leverages the homophily entropy to sample high-quality paths, streamlining the sampling process for optimal effectiveness.

\textbf{Adaptive walk number.}
    Echoing the success of the personalized walk length, the adaptive walk number also demonstrates a significant performance improvement over models that utilize a rigid walk number.
    It comes from our systematic evaluation of the information richness in sampling more informative walk sequences.

\subsection{Efficiency Analysis}

\begin{figure}[t]
\centering
\includegraphics[width=\linewidth]{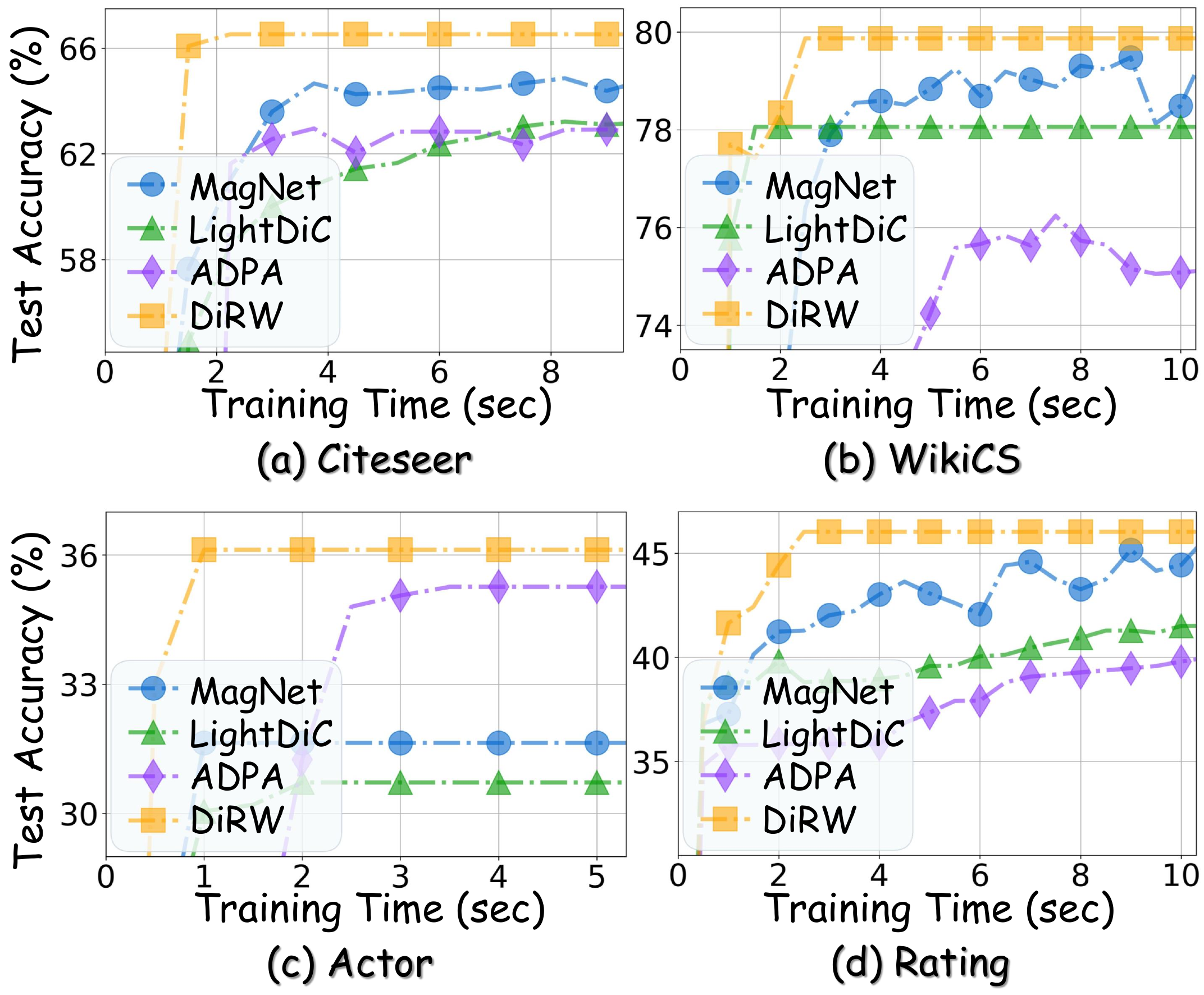}
\vspace{-6mm}
\caption{The Time-Accuracy performance comparison of DiRW and other DiGNNs.}
\vspace{-1mm}
\label{fig: effiency}
\end{figure}
    
    To address \textbf{Q3}, we conducted experiments to evaluate the efficiency of DiRW.
    We selected a spatial-based baseline, ADPA, and two spectral-based baselines, Magnet and LightDiC, to benchmark their performance and efficiency.
    The metric for efficiency was established based on the total time required to run each model five times, with the results graphically depicted in Fig.~\ref{fig: effiency}.

    From the visual representation, it is evident that the superiority of DiRW lies in both predictive accuracy and operational efficiency.
    Although optimal adaptive walk lengths and numbers vary between datasets, causing some fluctuations in computational time, DiRW achieves the best performance and efficiency due to its lightweight learning mechanism.
    ADPA is consistently the most time-consuming model, largely due to its intricate architecture, featuring hierarchical attention mechanisms and multiple convolution layers, which increases its computational and resource complexity.
    
    Notably, the sampling stage constitutes the primary computational bottleneck in DiRW's workflow, accounting for the majority of its running time.
    By adopting a pre-processing strategy analogous to PathNet, where we precompute and store sampled sequences, we can substantially optimize the training pipeline's efficiency.
    This preparatory approach synergizes effectively with DiRW's dedicated rapid learning mechanism, facilitating more accelerated training.
    
\begin{figure*}
\centering
\includegraphics[width=\linewidth]{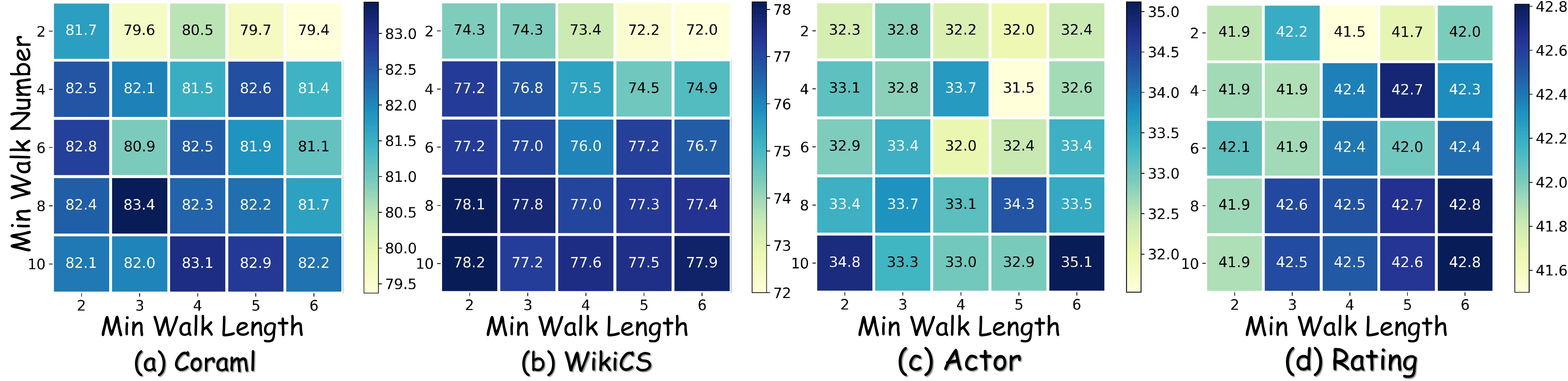}
\vspace{-6mm}
\caption{Sensitivity analysis of test accuracy to minimal walk length $l_{min}$ and minimal walk number $n_{min}$ on node classification.}
\vspace{-1mm}
\label{fig: heat}        
\end{figure*}

\begin{figure*}
\centering
\includegraphics[width=\linewidth]{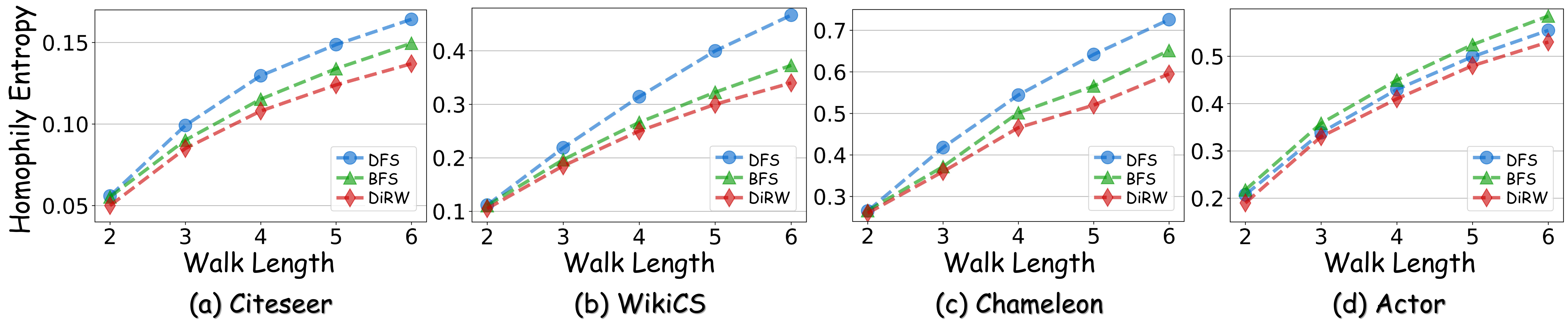}
\vspace{-6mm}
\caption{Path homophily entropy comparison of the optimized path sampler in DiRW and DFS/BFS in RAWGNN.}
\vspace{-1mm}
\label{fig: path}        
\end{figure*}

\subsection{Sensitivity Analysis}
    To evaluate the impact of hyperparameters on performance, we conducted an in-depth analysis of the minimum walk length $l_{min}$ and the minimum walk number $n_{min}$ in Fig.~\ref{fig: heat}.
    All other parameters were kept constant to isolate the effects of $l_{min}$ and $n_{min}$.

    Our analysis reveals a clear trend of improved performance as the minimum walk number $n_{min}$ increases.
    However, while more walks enhance performance, this cannot be pursued indefinitely due to the accompanying rise in time and space complexity.

    The impact of $l_{min}$ is less straightforward.
    There is no obvious performance improvement or decrease as $l_{min}$ increases.
    Notably, in homophilous digraphs (e.g., Coraml, WikiCS), shorter walk lengths yield better performance.
    Conversely, in heterophilous digraphs (e.g., Actor, Rating), longer walk lengths are linked to superior performance.
    This aligns with our earlier discussions on high-order homophily in heterophilous graphs, indicating that longer walks are crucial for capturing complex relationships within these structures.

\vspace{-2mm}

\subsection{Path Quality Analysis}
    We conducted an in-depth analysis comparing the path quality sampled by our optimized path sampler with those generated by the DFS and BFS utilized in RAWGNN, measured by the homophily entropy in Eq.~(\ref{eq: Homophily Entropy}).
    The experimental outcome is shown in Fig.~\ref{fig: path}.
    
    The results demonstrate that the paths sampled by DiRW consistently exhibit lower homophily entropy compared to those produced by DFS and BFS.
    This superiority is primarily attributed to the integrated consideration of both topology and feature in the walk probability, coupled with the adaptive walk length guided by homophily entropy.
    Furthermore, the DFS demonstrated the highest homophily entropy in most scenarios because BFS explores the immediate neighborhood, which aligns well with the homophily assumption.
    However, in scenarios characterized by strong heterophily (e.g., Actor), DFS outperforms BFS.
    DiRW achieves a harmonious balance between these two approaches, leveraging the strengths of both while mitigating their respective weaknesses.

\section{Conclusion and Future Work}
\label{sec: Conclusion}
    In this work, we emphasize the necessity of DiGNNs for modeling digraph-structured data and introduce the DiPathGNN mechanism to capture inherent homophilous information in digraphs.
    Our empirical analysis reveals the limitations of current PathGNNs, particularly their overlook of edge direction and coarse-grained sampling strategies.
    To overcome these challenges, we propose DiRW, a novel path-based digraph learning method characterized by its direction-aware path sampler, which is fine-tuned based on walk probability, length, and number.
    Additionally, DiRW utilizes a node-wise learnable path aggregator to create nuanced node representations.
    Experiments show that DiRW achieves state-of-the-art performance in node- and link-level tasks, particularly in heterophilous scenarios, providing an efficient and robust solution for digraph learning.
    Looking ahead, several promising directions for future research emerge.
    First, we plan to explore more efficient sampling strategies.
    Then, developing adaptive sampling methods would also improve model performance.
    Lastly, designing improved learning mechanisms to capture the complex relationships in digraphs may yield more accurate predictions.
    These future efforts will deepen our understanding of digraph learning and could lead to breakthroughs in modeling complex digraph-structured data.
    
\begin{acks}
    This work is supported by the NSFC Grants U2241211, 62427808, and U24A20255.
    Rong-Hua Li is the corresponding author.
\end{acks}

\section*{GenAI Usage Disclosure}
    We used ChatGPT exclusively for language polishing in this work.
    We take full responsibility for the integrity of the work’s intellectual content and its adherence to academic ethical standards.

\bibliographystyle{ACM-Reference-Format}
\balance
\bibliography{reference}

\end{document}